\pdfoutput=1

\documentclass[11pt]{article}

\usepackage[final]{acl}

\usepackage{times}
\usepackage{amssymb}
\usepackage{amsmath}
\usepackage{latexsym}

\usepackage[T1]{fontenc}

\usepackage[utf8]{inputenc}

\usepackage{microtype}

\usepackage{inconsolata}

\usepackage{graphicx}

\usepackage{microtype}
\usepackage{url}
\usepackage{booktabs}
\usepackage{multirow}
\usepackage{wrapfig, lipsum}
\usepackage{enumitem}
\usepackage[ruled]{algorithm2e}

\definecolor{redlinkcolor}{rgb}{0.79607843, 0.25098039, 0.25882353}
\definecolor{bluecitecolor}{rgb}{0,0.36,0.69}

\title{Temperature-Centric Investigation of Speculative Decoding with Knowledge Distillation}

\author{Siru Ouyang\textsuperscript{1\thanks{\ \ Work partially done during internship at Microsoft.}}, Shuohang Wang\textsuperscript{2}, Minhao Jiang\textsuperscript{1}, Ming Zhong\textsuperscript{1}\\
\textbf{Donghan Yu\textsuperscript{2}}, \textbf{Jiawei Han\textsuperscript{1}}, \textbf{Yelong Shen\textsuperscript{2}} \\
  \textsuperscript{1} University of Illinois Urbana-Champaign 
  \textsuperscript{2} Microsoft\\
  \normalsize\texttt{siruo2@illinois.edu} \\}

\begin{document}
\maketitle
\begin{abstract}
Speculative decoding stands as a pivotal technique to expedite inference in autoregressive (large) language models. This method employs a smaller \textit{draft} model to speculate a block of tokens, which the \textit{target} model then evaluates for acceptance. Despite a wealth of studies aimed at increasing the efficiency of speculative decoding, the influence of generation configurations on the decoding process remains poorly understood, especially concerning decoding temperatures.
This paper delves into the effects of decoding temperatures on speculative decoding’s efficacy. Beginning with knowledge distillation (KD), we first highlight the challenge of decoding at higher temperatures, and demonstrate KD in a consistent temperature setting could be a remedy. We also investigate the effects of out-of-domain testing sets with out-of-range temperatures. Building upon these findings, we take an initial step to further the speedup for speculative decoding, particularly in a high-temperature generation setting. Our work offers new insights into how generation configurations drastically affect the performance of speculative decoding, and underscores the need for developing methods that focus on diverse decoding configurations. Code is publically available at \texttt{\url{https://github.com/ozyyshr/TempSpec}}.
\end{abstract}

\section{Introduction}

\begin{figure}[h]
\centering
\includegraphics[width=0.48\textwidth]{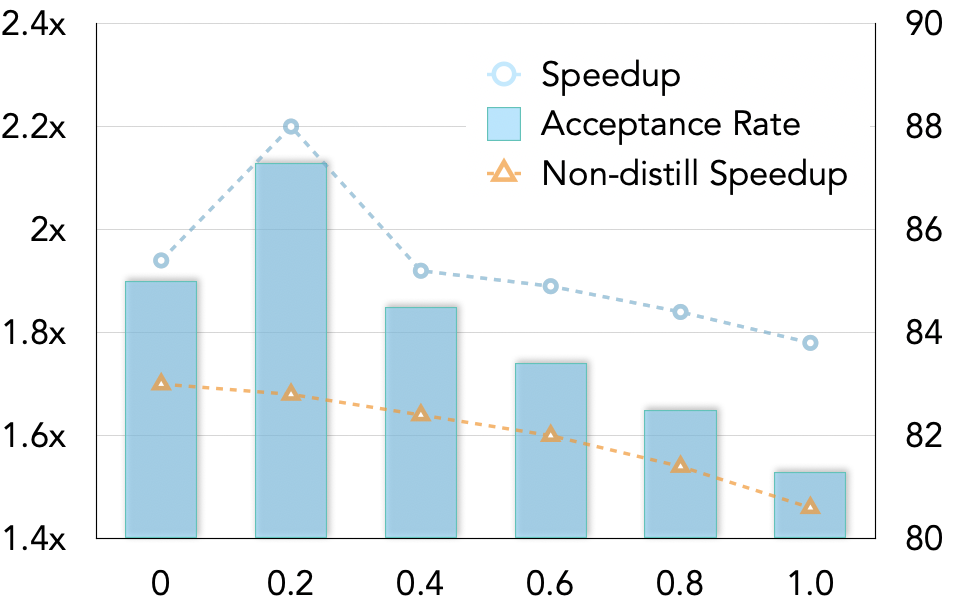}
\vspace{-5mm}
\caption{Speedup and acceptance rate (y-axises) for different decoding temperatures (x-axis) on Alpaca dataset. The draft model (Llama-68M) is distilled from Llama-2-13B-chat with data generated in $0.2$ temperature.}
\label{fig:intro}
\end{figure}

Large language models (LLMs) such as GPT-4~\citep{OpenAI_GPT4_2023}, Claude~\citep{DBLP:journals/corr/abs-2204-05862}, and LLaMA~\citep{touvron2023llama, touvron2023llama2} are revolutionizing the field of natural language processing (NLP) and machine learning (ML). While being powerful tools for various downstream tasks, LLMs' real-time deployment is still challenging due to the size and the inference cost~\citep{DBLP:journals/corr/abs-2211-05102}. Conversely, smaller models have less latency but lower generative quality. In a word, efficiency and accuracy form a trade-off. Inspired by this, speculative decoding~\citep{leviathan2023fast, chen2023accelerating} emerges as a promising \textit{token-level} solution to reduce the latency of generation for LLMs. Specifically, speculative decoding leverages smaller models as draft models to speculate successive candidate tokens for multiple inference steps with autoregressive generation, which are then verified with the target LLM in parallel through a \textit{single forward pass}. If a token fails to be accepted by the target LLM, all the consecutive tokens will be discarded, and the target LLM needs resampling for that rejected token.

Previous studies~\citep{xia2024unlocking} generally test speculative decoding in fixed generation configurations, with temperature sampling~\citep{Ackley1985ALA} being the default setting. Compared with other hyperparameters such as top-$k$~\citep{fan-etal-2018-hierarchical} in text generation, temperature has a dominating effect in re-estimating the distribution before top-$k$ sampling~\citep{Radford2018ImprovingLU}, balancing generation quality and diversity~\citep{Holtzman2020The}. However, previous works only test at a coarse-grained level, setting the temperature to binary extremes of either $0$ (greedy decoding) or $1.0$.  On the other hand, accelerating speculative decoding in various generation scenarios is important to better suit user needs in downstream tasks. To this end, this paper investigates from a \textbf{temperature-centric} perspective of speculative decoding for LLMs.

We focus on knowledge distillation (KD)~\citep{DBLP:journals/corr/HintonVD15} as the general investigation setting, which has been introduced as an intuitive and general solution to speculative decoding~\citep{zhou2023distillspec, liu2023online}. Particularly, KD aims to align the distributions of draft models better to that of target models. In this way, the \textit{acceptance rate} of candidate tokens generated by the draft model to the target model could be boosted. Our preliminary experiments in Figure~\ref{fig:intro} validate our motivation, highlighting the impacts brought by different temperatures for both decoding and KD stages. Notably, the speedup of the decoding processes increases and peaks at a decoding temperature of $0.2$ before declining as the temperature approaches $1$. The impact of temperature on speedup can reach a relative difference of around \textbf{30\%} ($\frac{2.23-1.72}{1.72}=29.7\%$), highlighting its importance. We also notice that KD relieves the degradation of speedup when temperature increases.

Overall, we explore the impact of temperature on speculative decoding with KD. Specifically, we address three pivotal research questions:

\begin{itemize}[leftmargin=*]
    \item \textbf{RQ1. What is the influence of temperature on speculative decoding’s efficacy in the context of KD?} To answer this question, we explore two key processes where the temperature is a critical factor in speculative decoding (\S~\ref{sec:background}). Utilizing the Llama series as the foundational model for both target and draft models, we train the draft model across a spectrum of training sets, each regulated by nuanced temperature settings, to assess and benchmark their performance (\S~\ref{sec:overall_investigation}). T5~\citep{DBLP:journals/jmlr/RaffelSRLNMZLL20} series is also explored for generalizability (\S~\ref{sec: model_combination}).
    \item \textbf{RQ2. Can the observed results extrapolate to out-of-domain datasets and out-of-range temperatures?} Building upon RQ1, we examine the adaptability of KD with temperature-specific configurations to \textit{out-of-domain} test sets derived from the training sets (\S~\ref{sec:ood_data}), and its performance with \textit{out-of-range} temperatures from those used during training (\S~\ref{sec:ood_temp}).
    \item \textbf{RQ3. How do we design an efficient recipe for enhancing speculative decoding in a temperature-centric manner?} Drawing from the insights of RQ1, we investigate various strategies for assembling training data with a temperature-aware approach (\S~\ref{sec:recipe}). Our goal is to amplify the performance of speculative decoding, particularly under conditions of elevated decoding temperatures.
    
\end{itemize}

The experiments are conducted on several commonly used public datasets. Our analysis offers a new perspective on understanding speculative decoding by applying fine-grained temperature controls, especially with KD. The key contributions and takeaways can be summarized as follows:

\begin{itemize}[leftmargin=*]
    \item We pinpoint \textit{temperature} as the key factor in the process of speculative decoding with KD. We empirically identify the most suitable setup, and find that temperature alignment between training and inference accelerates decoding significantly. 
    \item We explore both \textit{out-of-domain} test sets and \textit{out-of-range} decoding temperatures, and show the importance of token difficulties for out-of-domain sets and the ``U-curve'' phenomenon for out-of-range temperatures.
    \item Building upon our findings, we propose a simple yet effective data-centric strategy to boost the speedup for speculative decoding at high temperatures, and show that it can further improve the speedup of $12\%$-$20\%$. 
\end{itemize}

\section{Background}
\label{sec:background}


\subsection{Temperature in Decoding}

Temperature $\tau$ is an important hyperparameter in the configurations for decoding, which controls the randomness of predictions by scaling the logits before applying the softmax function during the text generation process~\cite{Ackley1985ALA}. It affects how the next word is chosen from the vocabulary:

\begin{equation}
    \mathbb{P}(t_k|t_{1:k-1}) = \frac{exp(l_k/\tau)}{\sum_i exp(l_i/\tau)},
\end{equation}
where $t_k$ and $l_k$ are the $k$-th token to predict and the corresponding logit. Lower temperatures will skew the distribution toward high-probability events, reducing the mass in the tail distribution to make the generation more focused and deterministic, and \textit{vice versa}. 

\subsection{Temperature in Knowledge Distillation}

The latency reduction actually depends on how \textit{aligned} the draft model and the LLM are. With better alignment comes lower rejection rates of tokens, thus higher acceleration speed. To make draft models better aligned with LLMs, KD is proposed as an intuitive yet effective solution~\citep{zhou2023distillspec, liu2023online}. In the KD process, the draft model $\mathcal{M}_d$ acts as the student, and the target model $\mathcal{M}_t$ serves as the teacher. We consider the two KD paradigms, \textit{online} and \textit{offline} distillation~\citep{zhong2024seeking}, in our investigation. Note that this paper focuses on lossless speculative decoding and the detailed algorithm for KD can be found in Appendix~\ref{app: kd_algorithm}.

During the KD process, the effect of temperature is mostly brought by the process of training data ($\mathcal{G}$) generation, which is contrastive to the temperature used in loss functions\footnote{In our investigation, the temperature in loss functions is always set to $1.0$ following previous works~\cite{chang2023kl}.}. Temperature guides the training data generation from the teacher model for offline data inference. Similar to offline distillation, the student model is asked to generate on-policy training data with temperature being the controlling factor in online distillation~\cite{agarwal2023gkd}. 

\paragraph{Offline Distillation}
We use SeqKD~\citep{kim-rush-2016-sequence} as the representative technique for offline distillation. It is a \textit{black-box} style framework, where only the teacher-generated texts are accessible. Training data are first generated by teacher $\mathcal{M}_t$ with decoding temperature $\tau$: 

\begin{equation}
\begin{split}
    y_i &= \mathcal{M}_t(x_i;\tau)\\
    \mathcal{G} &= \{(x_i, y_i)\mid i=1,2,...,n\}
\end{split}
\end{equation}
\noindent where $(x_i, y_i)$ denotes the input-output pair. The collected data are then used to train the student $\mathcal{M}_d^\theta$ parameterized by $\theta$:

\begin{equation}
\theta^* = \arg\min_{\theta} \sum_{(x_i, y_i) \in \mathcal{G}} \mathcal{L}(\mathcal{M}_d^\theta(x_i), y_i)
\end{equation}
\noindent The student model $\mathcal{M}_d^\theta$ is trained to minimize this loss, effectively learning to mimic the teacher’s behavior.

\paragraph{Online Distillation} In this setting, we assume \textit{white-box} access to both target and draft models, i.e., we can obtain the token-level distributions. Online distillation to the draft models seeks to minimize the divergence between the soft logits of teacher and student distributions over a training set, by using online data generated by $\mathcal{M}_d$:

$$\theta:= \arg \min \mathbb{E}_{(x,y)\sim \mathcal{G}}[D(\mathcal{M}_t||\mathcal{M}_d^\theta(y|x;\tau; \lambda))],$$

\noindent $\mathcal{D}$ measures the distance of two distributions and we use the default \textit{forward} Kullback-Leibler divergence (FKL) in our experiments. $\tau$ and $\lambda$ control the generation temperature and data fractions of the student model, respectively. Table~\ref{kd_settings} summarizes the setting of offline and online distillation.

\begin{table}
\centering
\setlength{\belowcaptionskip}{1 pt}
\small
\setlength{\tabcolsep}{1 pt}
\begin{tabular}{lcc}
    \toprule
    Setting    & Divergence ($\mathcal{D}$) & Training Data ($\mathcal{G}$)\\
    \midrule
    Offline Distill &FKL&Data generated by $\mathcal{M}_t$ offline \\
    \multirow{2}*{Online Distill} & \multirow{2}*{FKL} & Fixed dataset + Online data \\
    ~&~& generated by $\mathcal{M}_d$\\
    \bottomrule
  \end{tabular}
  \caption{Comparison of two settings for offline distillation and online distillation.}\label{kd_settings}
\vspace{-5mm}
\end{table}

\section{Experimental Setup}

This section outlines the detailed experimental setup, including model architecture, dataset selection, and evaluation metric employed for knowledge distillation (KD) and decoding phases. Further details on implementation, including hyperparameter configurations and computation timeframes, are provided in Appendix~\ref{app: exp}.

\subsection{Models and Datasets}
\paragraph{Models} In our experiments, we follow the settings of previous works~\citep{liu2023online, miao2023specinfer} and employ the Llama~\citep{touvron2023llama, touvron2023llama2} series as model architectures, a publicly available and prevalently used LLM family. Specifically, we select the instruction-tuned Llama-2-13B-chat~\footnote{\url{https://huggingface.co/meta-llama/Llama-2-13b-chat-hf}} as the target model, and Llama-68M~\footnote{\url{https://huggingface.co/JackFram/llama-68m}} as the draft model. The pre-trained model parameters for both models used are accessible via HuggingFace. 
\paragraph{Datasets} We focus on the general task of text generation with instructions. We use the Alpaca~\citep{alpaca} dataset as our fixed dataset following~\citep{miao2023specinfer}. The original Alpaca collection contains $52$k samples in the format of instruction-input-output triples, and we take $51k$ as the training set for KD. The rest of the $1k$ samples are left as \textit{in-domain} testing set. For offline distillation, we employ vLLM~\citep{kwon2023efficient} first to generate responses for each sample in the fixed dataset using the teacher model $\mathcal{M}_t$. The generated responses are paired with the original input as the training data for offline distillation. For online distillation, we use a half-fixed dataset with another half-on-policy data generated by student model $\mathcal{M}_d.$ All data generated by either $\mathcal{M}_t$ or $\mathcal{M}_d$ is based on temperature sampling with temperature $\tau$ in $[0,1]$ of interval $0.1$. That being said, we have a total of $11$ configurations of data generation in the KD process, which results in $22$ draft models for testing for both offline and online distillation settings. Apart from the $1k$ samples from Alpaca as the \textit{in-domain} set, we also use the GSM8K~\citep{cobbe2021gsm8k} test set containing $1.28k$\footnote{The original test set of GSM8K contains $1.32k$ samples, we filter out samples that exceed the context length of the draft model.} samples as the \textit{out-of-domain} set.

\subsection{Evaluation}

\paragraph{Metrics}
Following previous works~\citep{leviathan2023fast, miao2023specinfer, zhou2023distillspec}, we measure the empirical acceptance rate $\alpha$, and relative wall time (latency) improvement $\gamma$. $\alpha$ serves as the measure of how closely $\mathcal{M}_d$ approximates $\mathcal{M}_t$, and directly influences $\gamma$. In our implementations, we adapt the code from HuggingFace assisted decoding~\footnote{\url{https://huggingface.co/blog/assisted-generation}} and count the numbers of tokens generated by $\mathcal{M}_d$ and tokens accepted by $\mathcal{M}_t$ for $\alpha$. Time for decoding is documented for $\gamma$.

All the decoding processes are conducted based on temperature sampling with temperature $\tau\in[0,1]$ spanning $0.2$. The batch size is set to $1$ by default. For statistical robustness, we decode each sample $5$ times and take the averaged number of $\alpha$ and $\gamma$ and the final results.

\paragraph{Platforms} 
The KD training was executed over eight V100 NVIDIA GPUs, each with 32GB memory. The decoding phase for all draft models was carried out on a single A100 40G NVIDIA GPU for the consistency of our conclusions. 


\section{Experiments and Analyses}

Our experiments and analyses are organized in the following workflow. We start with an overall investigation of temperature configurations for two KD settings for in-domain testing. Leveraging these observations, we further test these insights on out-of-domain datasets with out-of-range temperatures. Finally, we brought out a simple yet effective solution to further improve the performance of speculative decoding with higher decoding temperatures.

\begin{figure*}[t]
\centering
\includegraphics[width=1\textwidth]{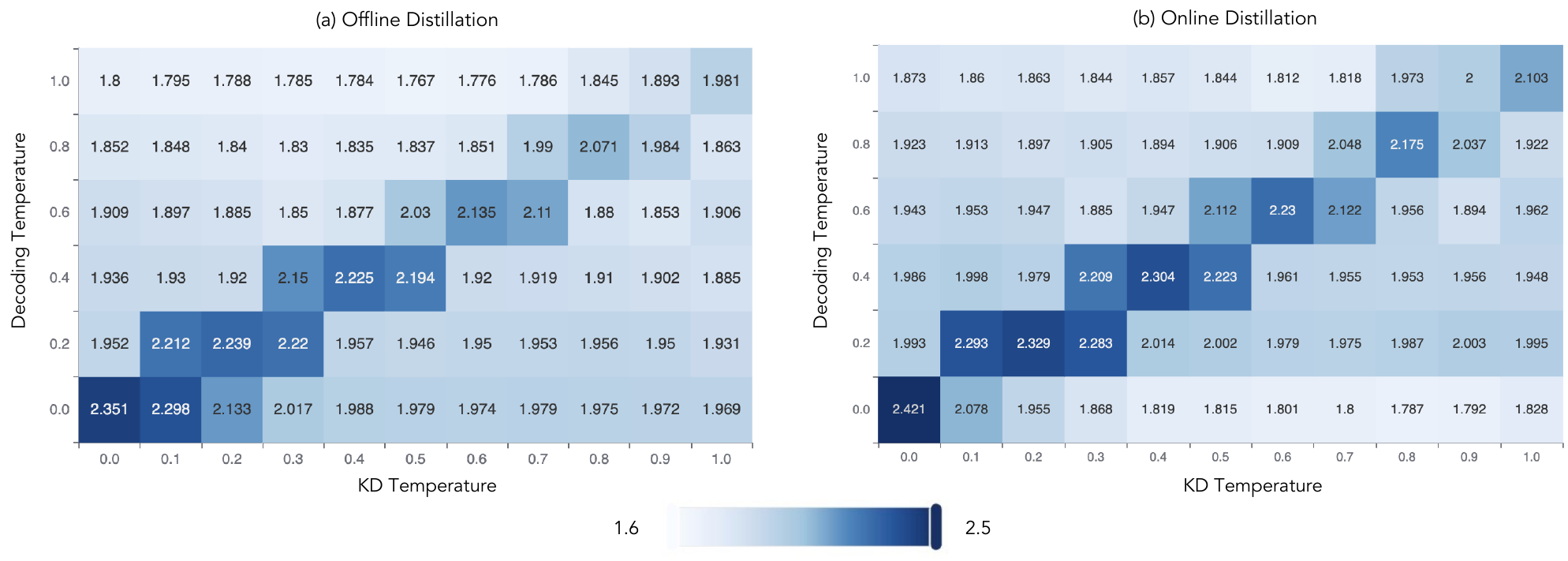}
\caption{Speedup for different decoding temperatures (y-axis) corresponding to different temperatures during KD (x-axis) for both (a) offline distillation and (b) online distillation for the testing of in-domain Alpaca set.}
\label{fig:overall}
\end{figure*}

\subsection{Overall Investigation}\label{sec:overall_investigation}
To quantify how temperature impacts the speculative decoding process, we plot the overall investigation results for both offline distillation and online distillation using $11$ KD models trained with different temperatures under $6$ decoding configurations in Figure~\ref{fig:overall} (a) and (b) respectively. We interpret the results in the following aspects. Additional analyses can be found in Appendix~\ref{app:additional_anal}.

\paragraph{Decoding at a high temperature is generally slower.}

\begin{figure}[t]
\centering

\includegraphics[width=0.45\textwidth]{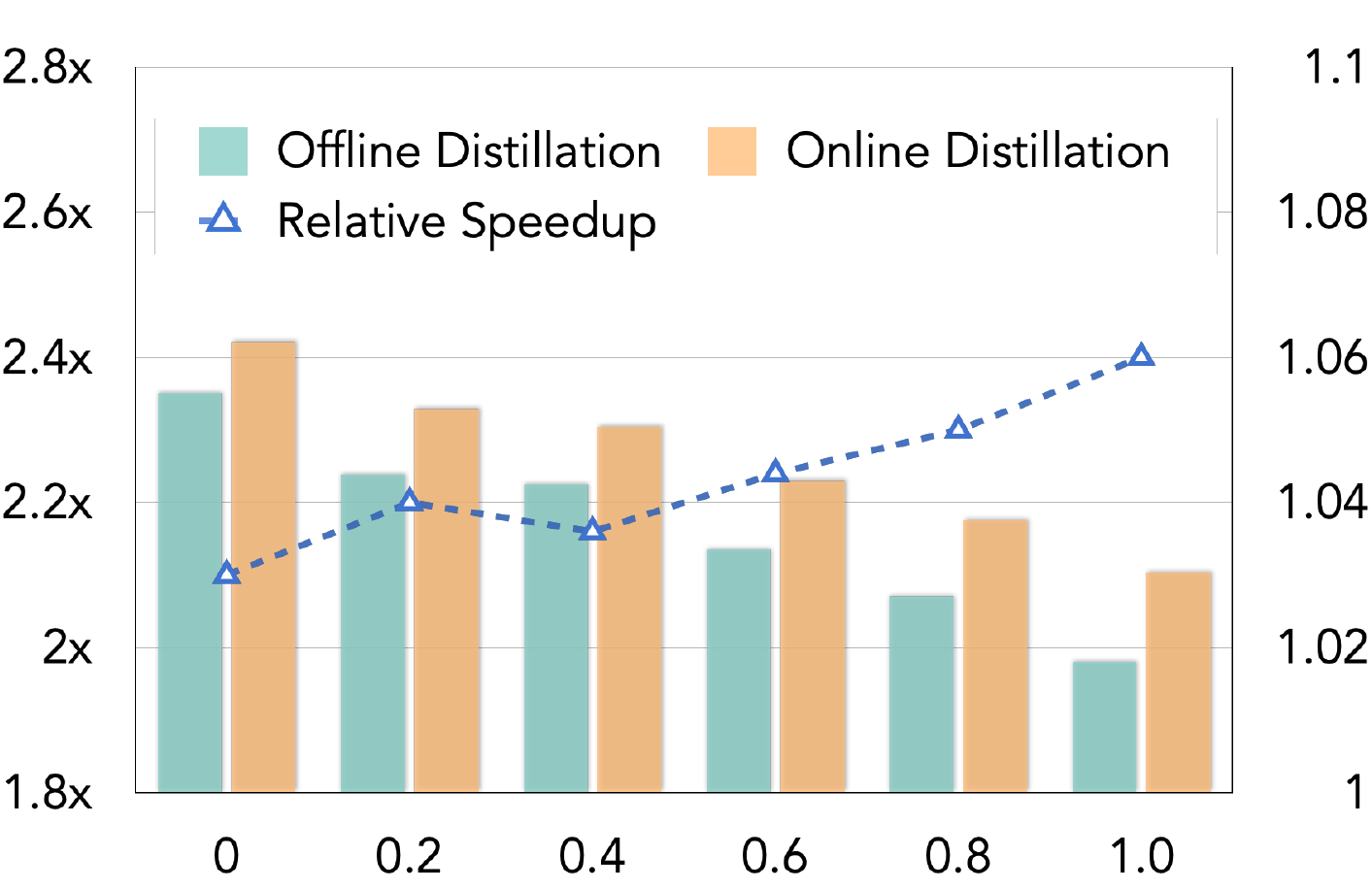}

\caption{Peak speedup brought by offline distillation and online distillation. The relative speedup for online distillation against offline distillation is depicted in dashed lines.}
\label{fig:relative_speedup}
\vspace{-5mm}
\end{figure}

First of all, we observe a consistent trend of diminishing speedup as the decoding temperature increases from $0$ to $1$. This trend corroborates the findings of previous studies, such as those by~\citet{xia2024unlocking}. Our analysis revealed that this phenomenon persists across all KD temperatures, affecting both offline distillation and online distillation processes. The effect was most pronounced when the KD temperature was set to $0$, leading to a relative speedup difference of $31\%$ and $29\%$ for offline distillation and online distillation, respectively. This is attributed to the increased computational complexity of the speculative sampling criterion at high temperatures, as demonstrated in prior research~\citep{gante2023assisted}. Thus, low temperatures are more likely to retain most of the latency benefits from generation via draft models. Additionally, we also observe that temperatures surrounding the peak values always lead to sub-optimal speedups. This is intuitive as the temperature can be seen as an approximate distribution measure. Apart from that, we find that higher temperatures in the surrounding ones usually lead to better results. For example, KD temperature at $0.7$ is better than $0.5$ when decoding at temperature $0.6$ even with the same temperature difference. This highlights another important factor, the diversity of data in KD, for the decoding process.

\paragraph{Using consistent temperatures for KD and decoding leads to better results.}

Our study reveals that configurations along the diagonals of Figure~\ref{fig:overall} typically yield the most accelerated decoding speeds. Grids outside the diagonals show pretty large differences with values on diagonals, with a peak relative difference of $24\%$. This verifies the effectiveness of KD at a consistent temperature. The speedup can be attributed to the alignment of probability distributions when the KD and decoding temperatures are nearly identical or perfectly match. We posit that this alignment facilitates a more efficient decoding process. Interestingly, as the decoding temperature increases, the speedup improvement resulting from this alignment diminishes. Specifically, for offline distillation, the relative improvement transitions from $31\%$ down to approximately $7\%$. Despite the challenges associated with accelerating speculative decoding at elevated temperatures, employing a uniform KD temperature for decoding --- particularly at $1.0$ --- proves to be more effective than using $0$. That being said, the upper right corner of Figure~\ref{fig:overall} is darker than the upper left corner. This finding further underscores the potential of KD as a remedy for alleviating the difficulty in decoding under high-temperature conditions. 

\paragraph{Online distillation is a better KD strategy for speculative decoding compared with offline distillation.}

Figure~\ref{fig:overall} illustrates that online distillation consistently outperforms offline distillation across a range of decoding temperatures. This is particularly evident at higher KD temperatures, where the student model benefits from softened probability distributions, allowing for a more nuanced understanding of the teacher’s distributions. For better observation, we also plot the peak speedup for every decoding temperature in Figure~\ref{fig:relative_speedup}, where the relative speedup of online distillation against offline distillation is in an increasing trend with higher temperatures. Additionally, we find that although online distillation surpasses offline distillation across multiple temperatures, the performance for online distillation at decoding temperature $0$ does not align with our expectations, especially with higher KD temperatures. Despite the alignment difference for binary temperature extremes between $1.0$ and $0$, the richer signal offered by online distillation could be another important factor since decoding at temperature $0$ usually entails hard labels.

\begin{figure*}[t]
\centering
\includegraphics[width=1.0\textwidth]{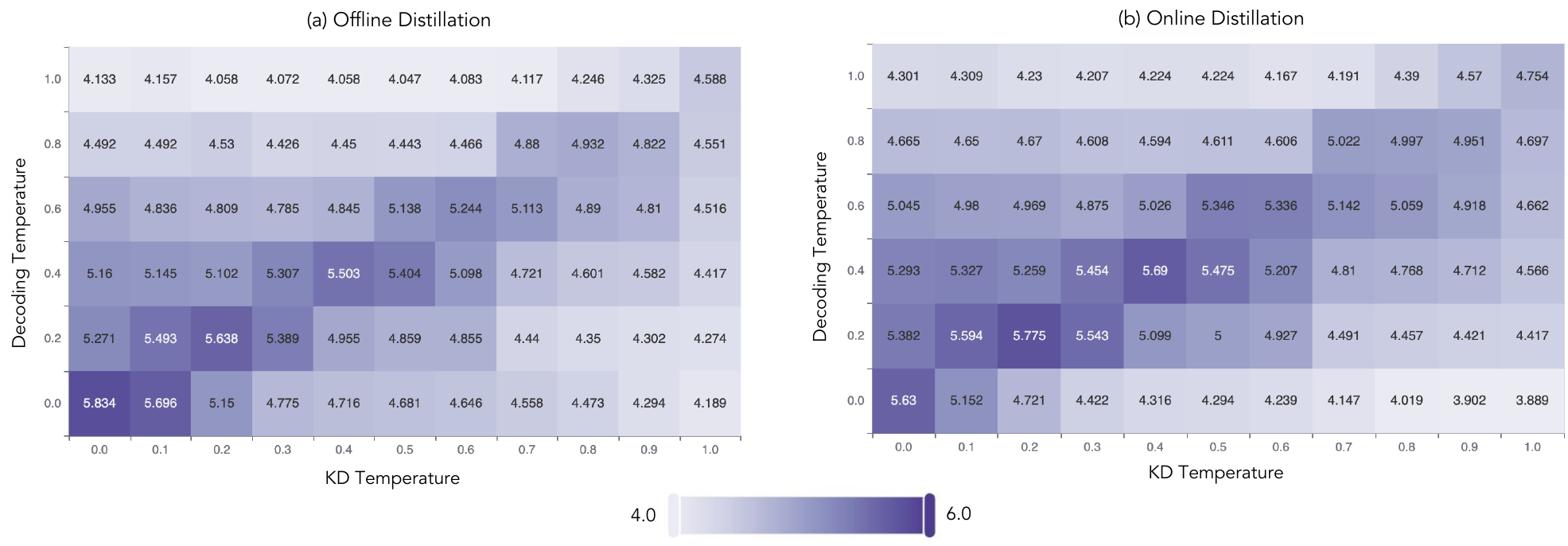}
\caption{Speedup for different decoding temperatures (y-axis) corresponding to different temperatures during KD (x-axis) for both (a) offline distillation and (b) online distillation for the testing of out-of-domain GSM8K set.}
\label{fig:gsm_overall}
\end{figure*}

\subsection{Evaluation on Out-of-domain Test Sets}\label{sec:ood_data}
 
To test whether our observations could be extended to out-of-domain datasets from training sets, we conduct experiments on GSM8K, a dataset focusing on multi-step graduate-school-level mathematical reasoning problems. It differs from the Alpaca training set that focuses more on general domains for everyday tasks. Results are shown in Figure~\ref{fig:gsm_overall}. 

\begin{figure}[t]
\centering
\includegraphics[width=0.43\textwidth]{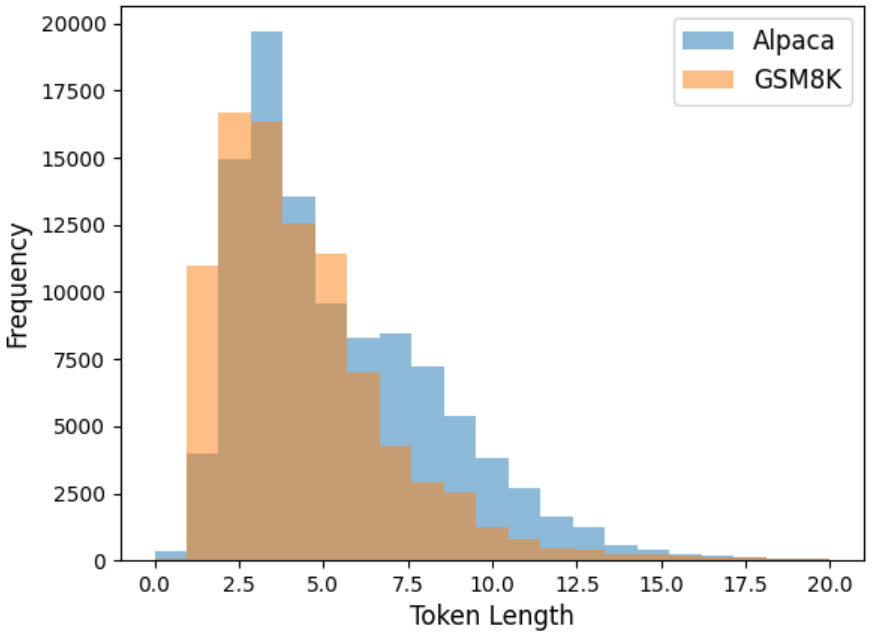}
\caption{The distribution of token length and the frequencies for both Alpaca and GSM8K test sets.}
\label{fig:token_distribution}
\vspace{-5mm}
\end{figure}

\textbf{Generally, the speedup brought by speculative decoding for GSM8K is much larger than that for the Alpaca set.} This could seem counter-intuitive for an out-of-domain testing set. One potential reason could be that the output for GSM8K consists of \textit{easier} tokens for the draft model to predict. Therefore, the acceptance rate is much higher for target models, which leads to a larger speedup. We found that the number of tokens generated for the Alpaca set ($18,716$) is much larger than that of GSM8K ($11,130$), around $68\%$ more than GSM8K, indicating the diversity in decoding processes. We also plot the distribution of token length for generation outputs in Figure~\ref{fig:token_distribution}. Intuitively, length can be seen as an approximate of the difficulty for that token. We observe that token length distribution for Alpaca is leaning towards longer tokens. This phenomenon sheds light on differentiating tokens of difficulties and designing corresponding strategies~\citep{shen2024learning} or employing Mixture-of-Experts structures~\citep{shazeer2017} at a token level.

The overall trend for GSM8K set at different decoding temperatures with KD settings is similar to Alpaca sets. Apart from this, we observe two other notably different phenomena. First of all, \textbf{the absolute differences in speedup across various temperatures for GSM8K are significantly larger than that for Alpaca.}  For example, with a KD temperature of $0$, the relative speedup difference achieved on the Alpaca set is around $30\%$ when the decoding temperature is set to $0$ and $1.0$, respectively. However, this value increases to $42\%$ for the GSM8K set. This pronounced variance indicates \textit{a stronger sensitivity} to the decoding temperature in the GSM8K set. Such sensitivity may be attributed to the nature of the mathematical reasoning tasks, which perhaps rely more critically on certain temperature thresholds to achieve optimal speculative decoding performance. Additionally, we find that \textbf{decoding at temperature $0$ with online distillation is particularly slow.} For one thing, the most aligned and fast choice of training under KD temperature $0$ does not yield the best speedup. Also, both offline distillation and online distillation do not yield strong performance at decoding temperature $0$. In contrast, offline distillation on the Alpaca set shows positive results.

\subsection{Evaluation on Out-of-range Decoding Temperatures}\label{sec:ood_temp}

In the previous experiments, we mainly focus on a traditionally recommended temperature range $[0,1]$ that makes LLMs respond in a human-acceptable way. To further understand the robustness and adaptability of our models, we have conducted additional experiments by evaluating them using out-of-range decoding temperatures. Specifically, we have expanded our evaluation to include decoding temperatures of $1.5$ and $2.0$, which are beyond the commonly used upper limit.

As illustrated in Table \ref{tab:ood_temp}, we observe several notable phenomena in the performance of both the Alpaca and GSM8K test sets when the decoding temperature is set to these higher values of $1.5$ and $2.0$. \textbf{First of all, we find a similar decreasing trend of speedup when the decoding temperature gets higher.} Specifically, we witness a relative difference of around $15\%$ of decoding at temperature $2.0$ compared with $1.5$. We also obtain the same observation where the speedup brought for offline distillation is larger than that for online distillation. However, the effect brought by different KD paradigms does not offset decoding temperatures. The effect of decoding temperatures tends to have different representations concerning datasets. Notably, GSM8K seems to have larger speedup differences for temperatures $1.5$ and $2.0$. This is because GSM8K has a higher speedup as baselines.

\textbf{Interestingly, the data reveals a distinctive \textit{U-curve} in the relationship between KD temperature and decoding speedup.} For instance, with the Alpaca test set at a decoding temperature of $1.5$, the speedup incrementally declines from $1.52$x at KD temperature $0$ to $1.45$x at KD temperature $0.4$, before ascending back to $1.58$x at KD temperature $1.0$. For one thing, increasing data diversity during KD training still helps for out-of-range and higher decoding temperatures, which might be caused by the somewhat approaching distributions with target models. However, speedup with KD temperature $0$ suggests that generation with fixed configurations holds a special meaning, 
potentially due to the alignment of distributions between the student and teacher models at this initial point. 

\begin{table*}[t]

\centering\centering\setlength{\tabcolsep}{4.9pt}
\small
\begin{tabular}{lcccccccccccc}
\toprule
\multirow{2}{*}{KD Temp.} &
\multicolumn{6}{c}{Offline Distillation} & \multicolumn{6}{c}{Online Distillation}\\
\cmidrule(lr){2-7}
\cmidrule(lr){8-13}
& \textit{0} & \textit{0.2}& \textit{0.4}& \textit{0.6}& \textit{0.8}& \textit{1.0}  & \textit{0} & \textit{0.2}& \textit{0.4}& \textit{0.6}& \textit{0.8}& \textit{1.0} \\
\midrule
\textbf{Alpaca test set} \\
w/ decoding temp. 1.5 &1.58x&1.55x&1.53x&1.52x&1.56x&1.60x&1.52x&1.49x&1.45x&1.50x&1.53x&1.58x\\
w/ decoding temp. 2.0&1.27x&1.25x&1.23x&1.26x&1.30x&1.35 x&1.22x&1.19x&1.16x&1.21x&1.23x&1.27x \\
\midrule
\textbf{GSM8K test set} \\
w/ decoding temp. 1.5&3.50x&3.48x&3.44x&3.47x&3.52x&3.59x&3.41x&3.36x&3.30x&3.34x&3.42x&3.48x \\
w/ decoding temp. 2.0&3.11x&3.09x&3.07x&3.04x&3.05x&3.07x&3.02x&2.93x&2.90x&2.92x&2.96x& 3.03x\\
\bottomrule
\end{tabular}
\caption{Results with out-of-range decoding temperatures on two KD settings with Alpaca and GSM8K test set.}\label{tab:ood_temp}
\vspace{-5mm}
\end{table*}

\subsection{Evaluation with Different Model Combinations}\label{sec: model_combination}

\begin{figure}[t]
\centering
\includegraphics[width=0.48\textwidth]{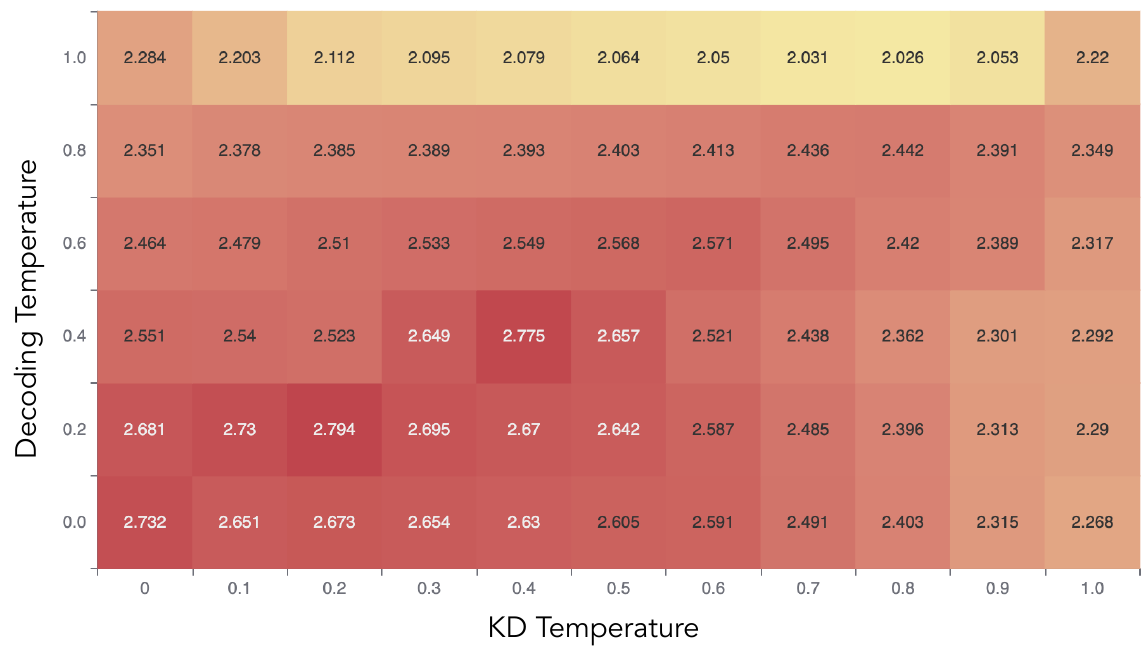}
\caption{Speedup for different decoding temperatures (y-axis) corresponding to different temperatures during KD (x-axis) for online distillation with T5-XL and T5-small on the testing of in-domain Alpaca set.}
\label{fig:t5_res}
\vspace{-3mm}
\end{figure}

To make our results more generalizable, we conduct experiments with an additional model family, T5~\citep{DBLP:journals/jmlr/RaffelSRLNMZLL20}. Specifically, we choose T5-XL (3B)~\footnote{\url{https://huggingface.co/google-t5/t5-3b}} and T5-small (60M)\footnote{\url{https://huggingface.co/google-t5/t5-small}} as the target-draft model pair for experiments. The rationale behind this choice is that (i) T5 stands for the encoder-decoder model family which is quite different from the Llama series, and (ii) T5 is commonly explored in previous works~\citep{DBLP:conf/icml/LiW0024}. Figure~\ref{fig:t5_res} presents the results. 

We can see that the general trend aligns with the previous conclusions in \S\ref{sec:overall_investigation}, i.e., decoding efficiency degraded when temperature decoding temperature goes high, while consistent temperatures lead to better speedups. Interestingly, the overall speedup for the T5 series exceeds that of the Llama series under the same distillation and testing conditions. We attribute this improvement to the more closely aligned model sizes of the target-draft pairs, with 3B-60M (T5) compared to 13B-68M (Llama). Additionally, the draft model for the T5 series was officially released by its original development team, whereas the Llama-68M model was trained by the open-source community, potentially introducing some discrepancies in the pre-training corpora. This discrepancy could be another potential reason for the lower speedups of the Llama series.

\subsection{Temperature-aware Recipe for Speculative Decoding}\label{sec:recipe}

In our prior investigations (as detailed in \S~\ref{sec:overall_investigation}), we establish that decoding at higher temperatures presents challenges. However, we also discover that KD can act as a promising remedy when training models under consistent temperature conditions. In this section, we propose a temperature-aware recipe for speculative decoding inspired by~\citet{chang2023kl}. Our approach employs a simple and intuitive \textit{data-centric composition} strategy, which represents an initial step toward enhancing decoding speed.

Specifically, we first manually identify the top-$k$ best-performing KD temperatures for the target decoding temperature from Figure~\ref{fig:overall} motivated by the following: (i) Values that approximate the best-performing temperature tend to align more with the target model's distribution; (ii) Diversity in training data for KD further boosts the performance. The selected temperature values are then used for KD in both settings for generation with teacher model and online student model generation. The detailed temperature configurations and experiment results are shown in Table~\ref{tab:recipe}. 

The composition data for KD are all chosen from the generation of the surrounding peak temperatures. On both Alpaca and GSM8K sets, we observe huge improvements in speedup, achieving an increase of $12\%$-$20\%$. 
Interestingly, a decoding temperature of $0.8$ with composition yields higher speedups than the higher temperature of $1.0$, suggesting that the influence brought by compositional data generation can fully make up for the slow speed when decoding at high temperatures. For the GSM8K dataset, similar trends are observed with even greater speedup values. For instance, with offline distillation and a KD temperature set of $\{0.9,0.8,0.7\}$, we achieve the highest reported speedup of $5.62×$ with an impressive acceptance rate of $89.5\%$.
Additionally, the observed differences in speedup gains between offline distillation and online distillation methods indicate that the former may be more amenable to training data composition strategies. These strategies, which leverage a set of temperatures rather than a single temperature, introduce a more nuanced control over the generated data's variability and quality. This granularity appears to be particularly beneficial for offline distillation, potentially due to the method's intrinsic reliance on the data itself as the primary source of knowledge transfer, which is well aligned with the black-box offline distillation.

\begin{table*}[t]

\centering\centering\setlength{\tabcolsep}{7.5pt}
\small

\begin{tabular}{lccccccc}
\toprule
\multirow{2}{*}{Methods} &Decoding&\multicolumn{2}{c}{KD temp.}&
\multicolumn{2}{c}{Alpaca} & \multicolumn{2}{c}{GSM8K}\\
\cmidrule(lr){3-4}
\cmidrule(lr){5-6}
\cmidrule(lr){7-8}
&temp.&Alpaca&GSM8K& \textit{Acc. Rate} & \textit{Speedup}  &\textit{Acc. Rate} & \textit{Speedup}  \\
\midrule
Offline distillation&1.0&1.0&1.0&80.6&1.98x&86.1&4.59x\\
&0.8&0.8&0.8&81.9&2.07x&87.3&4.93x\\
\quad w/ composition&1.0&\{1.0, 0.9, 0.8\}&\{1.0, 0.9, 0.8\}&83.0&2.23x&88.7&5.28x\\
&0.8&\{0.9, 0.8, 0.7\}&\{0.9, 0.8, 0.7\}&83.6&\textbf{2.34x}&89.5&\textbf{5.62x}\\
\midrule
Online distillation&1.0&1.0&1.0&82.2&2.10x&87.1&4.75x\\
&0.8&0.8&0.8&82.6&2.18x&87.9&5.00x\\
\quad w/ composition&1.0&\{1.0, 0.9, 0.8\}&\{1.0, 0.9, 0.8\}&83.5&2.27x&88.5&5.20x\\
&0.8&\{0.9, 0.8, 0.7\}&\{0.9, 0.8, 0.7\}&83.7&\textbf{2.33x}&88.9&\textbf{5.41x}\\
\bottomrule
\end{tabular}
\caption{Performance with data composition on two KD settings. Acceptance rate and speedup are reported for both in-domain and out-of-domain datasets.}\label{tab:recipe}
\vspace{-5mm}
\end{table*}
\section{Related Work}
\label{related_works}

\paragraph{Speculative Decoding} The sequential decoding strategy that is prevalently used in autoregressive Transformers~\citep{NIPS2017_3f5ee243} brings latency in real-world servings. To reduce the latency and accelerate decoding speed, the idea of parallel decoding was initially explored in various works~\citep{DBLP:conf/nips/SternSU18, ghazvininejad-etal-2019-mask}, with strict constraints and deviated distributions. 
Speculative decoding~\citep{leviathan2023fast, chen2023accelerating} brings success in reducing the inference latency of LLMs, some recent works~\citep{xia2024unlocking} have attempted to further improve speculative decoding by reducing the rejection rate of candidate tokens. 
Specifically, Predictive Pipeline Decoding~\citep{yang2023predictive} was proposed at first to incorporate early exit~\citep{schuster2022confident} into the decoding process. 
Later 
Another line of work is to leverage the target model for the self-drafting process, such as Draft\&Verify~\citep{zhang2023draft}, Medusa~\citep{cai2024medusa}, and Speed~\citep{hooper2023speed}. Tree attention is also explored, where multiple candidates during drafting are taken into consideration~\citep{miao2023specinfer}. 
Cascaded drafting process~\citep{spector2023accelerating, chen2023cascade} is invented to reduce drafting latency, alongside parallel drafting techniques~\citep{monea2023pass, xiao2024parallelspec} that attempt to mitigate the high latency issue of drafting. 
Specialized speculative decoding frameworks focusing on long-context~\citep{magicdec, sun2024triforce} and retrieval-augmented generation~\citep{rest, retreive_spec} have been proposed to align with task-specific characteristics.
However, almost all of the previous works only investigate the coarse-grained effect brought by generation configurations, such as temperature. For example, CSDecoding~\citep{chen2023cascade} and SpecInfer~\citep{miao2023specinfer} only explore greedy decoding for testing. Our work mostly relates to the work that leverages knowledge distillation~\citep{zhou2023distillspec, liu2023online}, with a focus on temperature-centric investigation for instruction-tuned KD draft models.

\paragraph{Knowledge Distillation for LLMs} Knowledge distillation (KD)~\citep{DBLP:journals/corr/HintonVD15} is a widely used model compression technique, aiming at training a student model with the guidance of a teacher model~\citep{gou2021knowledge}. The student model emulates the teacher models' behavior on downstream tasks. Standard KD methods are approximately minimizing the generation distribution of the student and the teacher. This is achieved by using the teacher's output at each time step as supervision~\citep{sanh2019distilbert} or direct training on the teacher's generated texts~\citep{kim-rush-2016-sequence}. With the emergence of LLMs, more techniques were invented for KD of LLMs, such as using reversed KL Divergence~\citep{gu2024minillm} or other variants of KLD~\citep{agarwal2023gkd, wen-etal-2023-f}. 
In this work, since we are targeting temperature-centric investigation of KD for speculative decoding, we only explore the two standard KD settings, i.e., black-box SeqKD~\cite{kim-rush-2016-sequence}, and online data generation that targets better KD for LLMs~\citep{agarwal2023gkd}.

\section{Conclusion}

In this paper, we have presented a comprehensive investigation into the impact of \textit{temperature} on speculative decoding, particularly within the context of knowledge distillation (KD), for large language models (LLMs). Through a series of meticulous experiments utilizing the Llama series as both target and draft models, we have explored the nuanced interplay between temperature settings during KD and their consequent effect on speculative decoding's efficiency and efficacy. Apart from offering empirical findings, we also propose a practical strategy to enhance speculative decoding's performance by leveraging temperature-centric training data assembly. By presenting this work, we aspire to facilitate future works on diverse generation configurations for speculative decoding, and exploring theoretical understanding of the multifaceted relations in between.

\section*{Limitations}
We discuss the limitations of this work in the following aspects:
\begin{enumerate}
    \item \textbf{Scope of the paper:} The factor of temperature for speculative decoding is an important aspect to investigate. While we investigated a general setting of knowledge distillation, we were not able to explore other settings due to limited computation resources.
    \item \textbf{Empirical analysis:} This study is an empirical investigation of the effect brought by different temperatures in speculative decoding. We interpret the conclusions and findings largely based on observations at hand, without solid theoretical foundations. Future works are encouraged to explore this direction.
    \item \textbf{Preliminary approach:} This study attempts to understand and accelerate speculative decoding at higher temperatures. We propose an empirical solution for data composition that has proven effective in our tests. However, our primary focus was not on developing comprehensive algorithms for speedup at higher temperatures. Further work could create more refined and mature solutions in this area.
\end{enumerate}

\section*{Acknowledgement}

Research was supported in part by US DARPA INCAS Program No. HR0011-21-C0165 and BRIES Program No. HR0011-24-3-0325, National Science Foundation IIS-19-56151, the Molecule Maker Lab Institute: An AI Research Institutes program supported by NSF under Award No. 2019897, and the Institute for Geospatial Understanding through an Integrative Discovery Environment (I-GUIDE) by NSF under Award No. 2118329. Any opinions, findings, conclusions, or recommendations expressed herein are those of the authors and do not necessarily represent the views, either expressed or implied, of DARPA or the U.S. Government.

\bibliography{custom}

\appendix

\section{KD Algorithm}\label{app: kd_algorithm}

In this section, we give the detailed algorithm of the KD training setting used in this paper.

\begin{algorithm}[htb]
\small
\caption{Online Distillation Algorithm}\label{Alg: knowledge_distillation}
\KwIn{Target model $\mathcal{M}_t$, Draft model $\mathcal{M}_d^\theta$, Data set containing (input, output) pairs}
\KwOut{Distilled draft model $\mathcal{M}_d^\theta$}
\textbf{Hyperparameters:} Data fraction from online generation $\lambda\in[0,1]$, Temperature $\tau\in[0,1]$, loss ratio $\gamma\in[0,1]$
\BlankLine
\For{$k$ in $0,...,n$}
{
    Generate a random value $\mu\in (0,1)$\\
    \If{$\mu\leq \lambda$}{sample inputs $x$ from $X$ and generate outputs $y$ by $\mathcal{M}_d^\theta(\cdot|x)$ with temperature $\tau$}
    \Else{sample inputs $x$ from $X$ and outputs $y$ from $Y$}
    
    Update $\theta$ to minimize $\mathcal{L}=\mathcal{L}_{lm}+\gamma\mathcal{D}(\mathcal{M}_t||\mathcal{M}_d^\theta(y|x))$
}
\end{algorithm}

\section{Implementation Details}\label{app: exp}

\paragraph{Data Formulation for Alpaca Dataset} For knowledge distillation, we instruction-tuned the model on the Alpaca dataset. Specifically, for each data sample in the dataset with triple ``instruction-input-output'', we use the following template to curate input for training:

If the elements in the triple are complete, we use the following template:

\texttt{Below is an instruction that describes a task, paired with an input that provides further context. Write a response that appropriately completes the request. \#\#\#Instruction:\{instruction\} \#\#\# Input:\{input\}\#\#\# Response:} 

If there is only ``instruction'' for the data sample without ``input'', the above template will be simplified as:

\texttt{
    Below is an instruction that describes a task. Write a response that appropriately completes the request.\#\#\# Instruction:\{instruction\}\#\#\# Response:}

\paragraph{Implementation Details for KD} For online distillation, we set the batch size to $8$, learning rate to 3e-5, maximum length of input to $512$. The training process continues for $30$ epochs with $200,000$ steps in total. It takes around $30$ hours to finish. For offline distillation, it takes $8$ hours to finish.

\paragraph{Implementation Details for Evaluation}
We set the maximum decoding length to $128$ due to the limit in A100 40G' GPU memory. Each evaluation corresponding to KD temperatures and decoding temperatures requires around $12h$ to run on the A100 GPU with batch size $1$.

\section{Detailed analysis for Section~\ref{sec:overall_investigation}}~\label{app:additional_anal}

\paragraph{Speedup is hard to get offset with longer KD steps.}

\begin{figure}[t]
\centering

\includegraphics[width=0.45\textwidth]{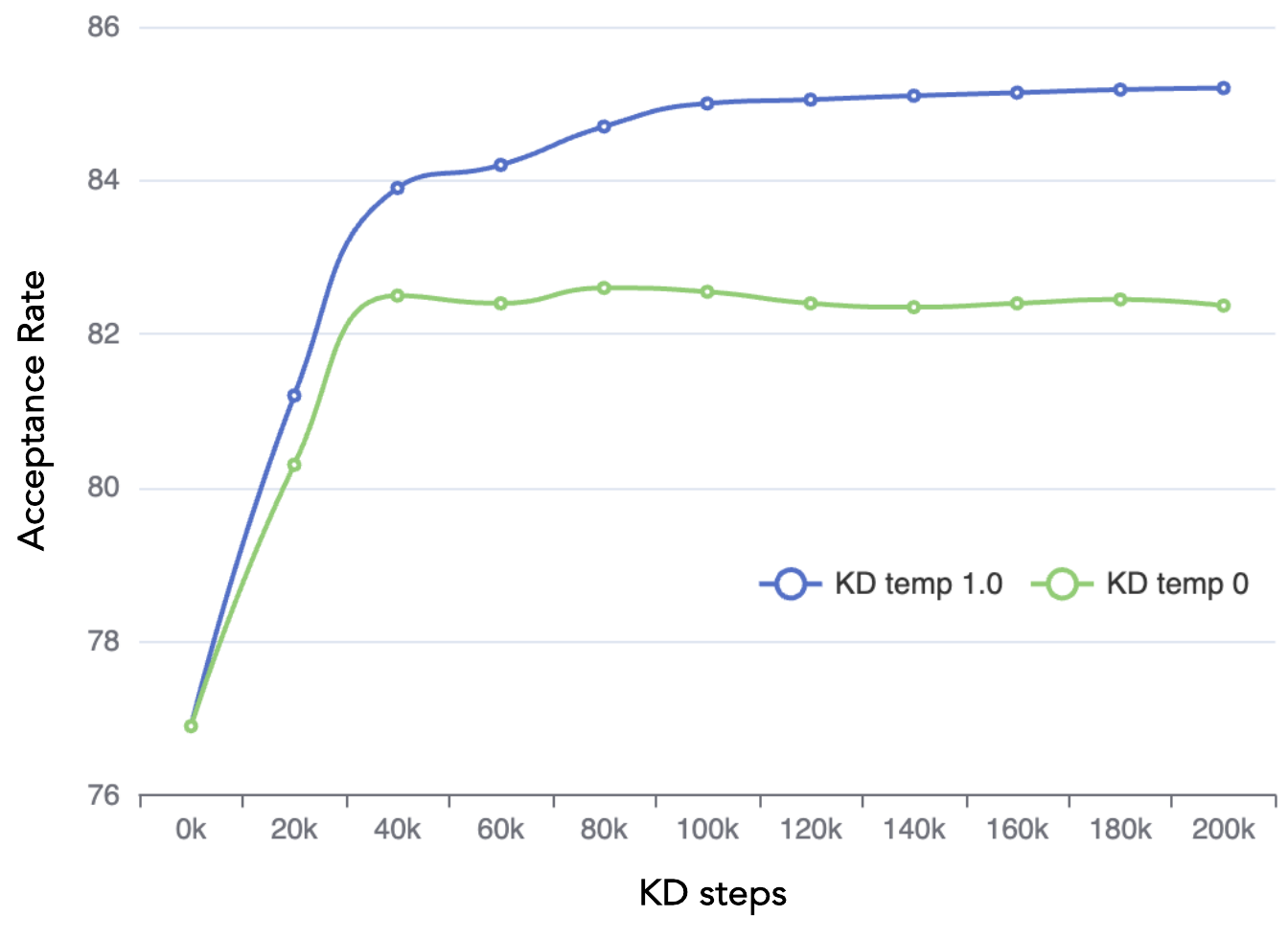}

\caption{Acceptance rate of different KD temperatures for decoding at temperature $1.0$ regarding KD steps on the Alpaca test set.}
\label{fig:kd_steps}

\end{figure}

According to our observation, the optimal performance is achieved when the decoding temperature and KD temperature align with each other. To further understand the improvement in speedup regarding the temperatures, we study the relation with KD steps in Figure~\ref{fig:kd_steps}.
We consider a rather extreme setting where the decoding temperature is set as $1.0$ with KD temperatures $0$ and $1.0$. During the initial stages of knowledge distillation, the two curves representing different temperature settings exhibit rapid growth and are relatively close to each other. As the KD process progresses, the curve with KD temperature $1.0$ diverges significantly from the other and the acceptance rate still steadily increases. As the KD process gradually approaches the end, the curve with KD temperature $1.0$ achieves higher speedup and continues to show an upward trend, whereas the other temperature curve plateaus with a lower acceptance rate.

\paragraph{Phenomenon of symmetric temperature configurations.} 
Intuitively, we might expect that distilling from a teacher with temperature $\tau_1$ and then using decoding temperature $\tau_2$ can behave similarly to distilling with temperature $\tau_2$ and decoding with temperature $\tau_1$.
This phenomenon could be referred to as diagonals (from upper left corner to lower right) in Figures~\ref{fig:overall}. We find that symmetric temperature settings do bring similar speedups. However, decoding at lower temperatures is still faster than at higher temperatures.

\end{document}